\documentclass[journal]{IEEEtran}

\usepackage{graphicx} 
\usepackage{booktabs}
\usepackage{amsmath}
\usepackage{mathrsfs}
\usepackage{amsfonts}
\usepackage{amssymb}
\usepackage{dblfloatfix}
\usepackage[caption=false,font=footnotesize]{subfig}
\usepackage{color}
\usepackage{enumitem}
\usepackage{amsthm}
\usepackage{algorithm,algorithmic}
\usepackage{multirow} 
\usepackage{array}
\usepackage{cite}

\begin{document}

\title{Deep Cross-Subject Mapping of Neural Activity}

\author{\IEEEauthorblockN{Marko~Angjelichinoski\IEEEauthorrefmark{1}, Bijan Pesaran\IEEEauthorrefmark{2}, and Vahid Tarokh\IEEEauthorrefmark{1}}\\
\IEEEauthorblockA{\IEEEauthorrefmark{1}Department of Electrical and Computer Engineering, Duke University}\\
\IEEEauthorblockA{\IEEEauthorrefmark{2}Center for Neural Science, New York University}
\thanks{This work was supported by the Army Research Office MURI Contract Number W911NF-16-1-0368 and DURIP Contract W911NF-21-1-0080.}
}

\maketitle

\begin{abstract}
\emph{Objective.} In this paper, we consider the problem of cross-subject decoding, where neural activity data collected from the prefrontal cortex of a given subject (destination) is used to decode motor intentions from the neural activity of a different subject (source).
\emph{Approach.} We cast the problem of neural activity mapping in a probabilistic framework where we adopt deep generative modelling. Our proposed algorithm uses deep conditional variational autoencoder to infer the representation of the neural activity of the source subject into an adequate feature space of the destination subject where neural decoding takes place.
\emph{Results.} We verify our approach on an experimental data set in which two macaque monkeys perform memory-guided visual saccades to one of eight target locations.
The results show a peak cross-subject decoding improvement of $8\%$ over subject-specific decoding.
\emph{Conclusion.} We demonstrate that a neural decoder trained on neural activity signals of one subject can be used to \textit{robustly} decode the motor intentions of a different subject with high reliability. This is achieved in spite of the non-stationary nature of neural activity signals and the subject-specific variations of the recording conditions. \emph{Significance.} The findings reported in this paper are an important step towards the development of cross-subject brain-computer that generalize well across a population.
\end{abstract}


\IEEEpeerreviewmaketitle

\section{Introduction}\label{sec:intro}

\subsection{Motivation}\label{sec_intro:motivation}

A key objective of neural engineering is to restore/supplement/enhance human capabilities by means of \emph{brain-computer interfaces (BCIs)} that generalize well across unseen subjects of a population \cite{rao_2013,TIWARI2018118}.
Example applications that motivate the development of cross-subject BCIs include \cite{Bensmaia:2014,MOXON201555}: 1) common clinical practices for treating neurological disorders such as epilepsy, Parkinson's disease, Alzheimer's disease and other debilitating conditions, 2) neural prosthetics, aiming to restore lost and/or chronically impaired motor functions, 3) public safety, and 4) tactical domain.

The implementation of reliable cross-subject BCIs is a notoriously challenging problem.
An important factor that contributes to the difficulty of cross-subject BCI arises from the \emph{non-stationary} nature of the neural activity signals, whose statistical properties vary dramatically even under slight changes of the recording conditions \cite{rao_2013,Jayaram_2016}.
As a result, BCI algorithms trained and optimized on data collected from a given subject, fail to perform reliably when directly applied to other subjects. 
Cross-subject BCI is further hindered by the {limited training data}, which is a typical circumstance because acquiring training data is expensive and time-consuming endeavour \cite{Waldert2016}.

The focus of this paper is on cross-subject BCIs in the context of \emph{neural decoding} of motor intentions where we consider, develop and evaluate solutions that address the above challenges.
Before proceeding to discuss the technical details, we first present a motivating example. 

\subsection{Running Example}\label{sec_intro:example}

\begin{figure}
    \centering
    \includegraphics[scale = 0.32]{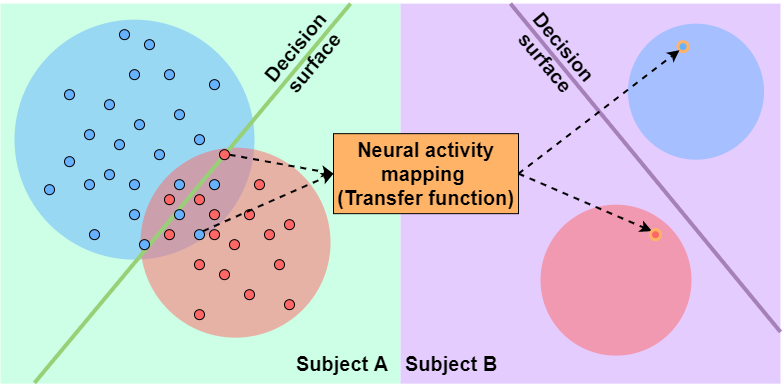}
    \caption{Schematic depiction of cross-subject neural decoding using single transfer function.}
    \label{fig_TL_cartoon}
\end{figure}

In the basic variant of \emph{cross-subject decoding}, see Fig.~\ref{fig_TL_cartoon}, data from two subjects, denoted as \emph{Subject A} (source) and \emph{Subject B} (destination), is available.
We assume that the subjects perform tasks using the same set of motor actions $\mathcal{K}=\{1,...,K\}$; typical examples of motor actions are arm reaches, head tilts, and eye movements to different directions.
The goal of cross-subject decoding is to infer the motor action that Subject A is performing using a classifier, also referred to as \emph{neural decoder}, trained on neural activity data collected from Subject B.
\footnote{In neural prostheses, Subject A can be understood to represent the subject with motor impairments, that has lost the ability to perform one/several motor action(s), whereas Subject B is the subject with normal motor functions.
In this scenario, the feature space representation of the brain activity of the disabled subject will be unbalanced and motor intentions related to impaired function(s) will be poorly decoded.
The objective, instead, is to use the neural decoder of the fully able subject to infer intended actions by the disabled subject with the hope of restoring the impaired functions.
The above ideas can be extended to other BCI applications, involving different objectives and/or number of subjects; this includes subject-specific applications that involve acquisition, manipulation and utilization of neuronal activity signals collected from different cortical sites/depths of the brain, such as, for instance, detecting, isolating and analyzing the causes of epileptic seizures in patients by mapping the neural activity from affected to non-affected areas of the brain.}

Let $X$ and $Y$ denote the random vectors representing the \emph{features} extracted from neural activity recordings of Subjects A and B, respectively,; both $X$ and $Y$ live in $\mathbb{R}^D$ where $D\geq 1$ is the dimension of the feature space.
Also, let $\delta_{\text{B}}(\cdot)$ denote the discriminant function of the neural decoder trained on Subject B training data; the objective of the cross-subject decoding problem can be formalized as
\begin{align}\label{eq:crosssubject_decode_general}
    \hat{k} = \arg\max_{k\in\mathcal{K}}\delta_{\text{B}}\left(X\right).
\end{align}
However, the non-stationary nature of neural activity signals implies that the class-conditional distributions, representing the same tasks in the feature spaces, differ across different subjects \cite{Jayaram_2016,Angjelichinoski_2020_CS}. 
As a result, a reliable neural decoder $\delta_{\text{B}}(\cdot)$ trained on Subject B will perform poorly if used \emph{directly} on Subject A; in most cases, the decoding performance is no better than a random chance.
To address this issue, we borrow ideas from transfer learning and we postulate that the feature space of subject B can be viewed as a \emph{functional transformation} of the feature space of subject A \cite{Angjelichinoski_2020_CS}: 
\begin{align}\label{eq:XY_mapping_unsupervised}
    Y = g\left(X\right)\quad\text{for all}\quad k\in\mathcal{K}.
\end{align}
Here, the feature vectors $X$ and $Y$ are drawn from the class-conditional distribution corresponding to the specific motor action $k\in\mathcal{K}$ while the mapping $g:\mathbb{R}^D\mapsto\mathbb{R}^D$ relates the feature space representations of the action set $\mathcal{K}$ in two different subjects; we refer to this map as a \emph{transfer function}.
Using the model \eqref{eq:XY_mapping_unsupervised}, we can revise the statement \eqref{eq:crosssubject_decode_general} as
\begin{align}\label{eq:crosssubject_decode_TL}
    \hat{k} = \arg\max_{k\in\mathcal{K}}\delta_{\text{B}}\left(g\left(X\right)\right).
\end{align}
Note that even though the transfer function can differ across tasks, in \eqref{eq:XY_mapping_unsupervised} we have removed the dependence of the transfer function on the task label since the labels of the test points are not known in testing time; instead, we aim to find a \emph{single} transfer function $g(\cdot)$, valid for \emph{all} tasks in $\mathcal{K}$.

\subsection{Contributions and Main Results}\label{sec_intro:contributions}

We propose an end-to-end cross-subject neural decoding system that 1) uses \emph{Pinsker's theorem} to extract relevant features from \emph{scarce} neural activity data \cite{Johnstone2012GaussianE,Banerjee1,Angjelichinoski2019,Angjelichinoski_2020_DJSNN}, and 2) applies \emph{nonlinear} cross-subject maps using deep conditional generative model, 
taking advantage of the ability of such models to capture complex, multimodal distributions \cite{CVAE_NIPS2015}.
We evaluate our approach on the problem of decoding eye movement goals from local field potentials (LFPs) collected in macaque lateral prefrontal cortex, while the subjects perform memory-guided visual saccades to targets placed at one of eight locations \cite{Markowitz18412}.
The results show that with training/testing data sets of size $1200/200$ trials, the subject-specific, i.e., local neural decoder trained on Monkey A data achieves decoding accuracy of $\approx 75\%$ on the test data; after applying cross-subject mapping, the test data from Monkey A can be decoded with average accuracy of $\approx 81\%$ using a neural decoder trained on data from Monkey B, marking a relative performance improvement of nearly $8\%$.
Even more so, our deep cross-subject neural mapping approach yields better performance than the benchmark performance established by \emph{linear} cross-subject mapping in \cite{Angjelichinoski_2020_CS}, by a relative margin of nearly $45\%$ over the same test data.

\subsection{Related Work}\label{sec_intro:relatedwork}

Domain adaptation methods are popular in non-invasive, electroencephalography (EEG)-based cross-subject BCIs \cite{Jayaram_2016,BCI_TL_AVA_2019,Zanini2018,Azab_2019,Kwon_deepConv_2019,Samek2013}; it should be noted that, in contrast with invasive BCIs, acquiring training data in non-invasive setups is relatively straight-forward procedure even with human subjects.
Few recent studies consider invasive BCIs \cite{Sussillo_2012,sussillo2016making,Pandarinath152884}; their attention has so far been mainly focused on subject-specific inter-session neural decoding, aiming to address the temporal variability of neuronal firing rates.
The work presented in \cite{Angjelichinoski_2020_CS} is, to the best of our knowledge, the most relevant to our problem as it addresses cross-subject neural decoding from LFP data and it verifies the solutions over the same experiment and data sets. 
The authors in \cite{Angjelichinoski_2020_CS} consider \emph{linear} neural mapping in the opposite direction: the training data is \emph{linearly} transferred from Subject B to Subject A feature space where a \emph{linear} classifier is trained.
This approach allows for supervised training of multiple transfer functions, one for each task in the action set; however, as reported in \cite{Angjelichinoski_2020_CS}, the cross-subject decoding performance is upper-bounded by the local, subject-specific performance of Subject A.

\section{Methods}\label{sec:sys_model}

A cross-subject neural decoding system consists of four main building blocks: 1) \emph{data acquisition} (i.e., brain recording), where the raw neural activity signals are collected, 2) \emph{feature extraction}, where the neural activity data is represented in (usually lower-dimensional) feature space, suitable for further processing, 3) \emph{cross-subject mapping}, where Subject A features are mapped to Subject B feature space using the transfer function, and 4) \emph{decoding}, where Subject A motor intentions are inferred using Subject B classifier. This section establishes the baseline for each of these blocks.

\subsection{Experimental Protocol and Data Acquisition}\label{sec_eval:protocol}

\begin{figure}
    \centering
    \includegraphics[scale = 0.3]{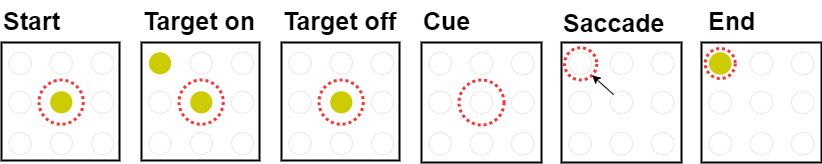}
    \caption{Experimental protocol: timeline of individual trials.}
    \label{exp_time}
\end{figure}

We begin with an overview of the experiment and the acquired data.
All procedures were approved by the NYU University Animal Welfare Committee (UAWC).

We study a classic experiment: memory-guided visual saccades to a target location \cite{Markowitz18412}.
Two adult macaque monkeys (\emph{M. mulatta}), referred to as Monkey A (Subject A) and Monkey B (Subject B)\footnote{Note that, in \cite{Markowitz18412,Angjelichinoski_2020_CS}, Monkey A and B are originally denoted as Monkey S and A, respectively.}, are trained to perform memory-guided visual saccades to one of eight target locations on a screen, see Fig.~\ref{exp_time}.
Individual trials are initiated by instructing the subject to fixate  a central visually-presented target. 
Once the subject maintains fixation for a baseline period, one of the eight peripheral visual targets (drawn uniformly at random from the corners and edge midpoints of a square centered on the central target) is turned on for the duration of $300$ milliseconds; each time, the target light is chosen \emph{independently} from previous selections.
The extinguishing of the peripheral target marks the beginning of the \emph{memory period} during which the subject must maintain fixation on the central target until it is extinguished; this event instructs the subject to saccade to the remembered location of the peripheral target.
The trial is completed successfully if the subject maintains the gaze within a small square window from the true target location.
We use only segments of neural activity recorded during the memory periods of successful trials; this epoch is especially interesting in memory-guided behaviors as the epoch presents information that determines the dynamics of the decision-making process and the subsequent motor response \cite{Markowitz18412}.

Recent advances have shown that \emph{local field potential (LFP)} signals present a viable alternative for designing invasive BCIs where the neural activity is recorded directly from brain tissue, via chronically implanted arrays of micro-electrodes \cite{Pesaran:2018}. LFPs refer to the potential of the extracellular currents surrounding individual neurons and, unlike the neuronal spiking activity, the LFP modality is more resilient to signal degradation \cite{Pesaran:2018}.
In the experiment considered in this paper, LFP activity was sampled at $\nu_S=1$ kHz and recorded using a microelectrode array with $N=32$ individually movable microelectrodes, semi-chronically implanted in a recording chamber placed over the lateral prefrontal cortex (PFC) \cite{Markowitz18412}.
The data set of Monkey B (i.e., Subject B) consists of 1400 trials and was collected at superficial cortical site, at a mean cortical depth of $\approx$ 1 millimeter (mm) from the cortical surface.
Monkey A (i.e., Subject A), on the other hand, is used as Subject A; as the experiment progressed, the vertical positions of individual electrodes in Monkey A were gradually advanced deeper while their horizontal coordinates remain unchanged, covering a range of cortical depths and yielding four data sets, each with 1400 trials, collected at mean depths of 1.4, 2.2, 2.8 and 3.5 mm below the surface of the PFC. 

\subsection{Feature Extraction}\label{sec_sysmodel:Pinsker}

The theory of non-parametric regression has been useful for extracting meaningful features that enable deep models to be trained reliably on limited data \cite{Banerjee1,Angjelichinoski2019,Angjelichinoski_2020_DJSNN}. We give a brief summary of the method. 

Let $\tilde{x}_{t},t=1,\hdots,T$ to denote the $t$-th LFP sample acquired from an arbitrary electrode in Subject A (analogous notation applies to Subject B).
We postulate that LFP activity consists of at least two components: (1) useful, information-carrying signal that determines the decision-making process, given by the function $x$, and (2) noise-like waveform $\sigma w$ representing the remaining part of the LFP which does not contribute to the dynamics of the decision-making and is modelled as independent and identically distributed (i.i.d.) Gaussian noise. These two components summate according to the following model:
\begin{align}\label{eq:lfp_model}
    \tilde{x}_t = x_t + \sigma w_t,\quad w_t\sim\mathcal{N}(0,1),\quad t=0,\hdots,T-1,
\end{align}
where $x_t = x(t \nu_S)$ and $w_t$ are the corresponding discrete versions of $x$ and $w$ respectively.
We do not assume parametric model for $x$; we only assume that the model lives in a space of smooth functions \cite{Johnstone2012GaussianE}.
Each different task $k\in\mathcal{K}$ yields different representation in the function space.
In addition, the signal $x$ will be also different across repeated trials due to variety of neurological reasons. 
Hence, it is accurate to say that each specific task $k\in\mathcal{K}$ forms a class of functions in the function space.

A desirable property of the neural decoder is to be consistent which can be guaranteed by taking the worst-case miss-classification probability to zero.
This motivates the use of \emph{minimax-optimal} function estimators \cite{Johnstone2012GaussianE}.
The theory of Gaussian sequence models provides a framework for designing finite-dimensional representations of the minimax-optimal function estimators.
We first project the LFP model \eqref{eq:lfp_model} onto an orthonormal set of functions, such as the Fourier basis functions to obtain the following sequence space representation:
\begin{align}\label{eq:GaussianSequence}
    \tilde{X}_l = X_l + \frac{\sigma}{\sqrt{T}} W_l,\quad W_l\sim\mathcal{N}(0,1),\quad l=1,2,\hdots,
\end{align}
where, $\tilde{X}_l$, $X_l$ and $W_l$ are the projections of the vectors $(\tilde{x}_0,\hdots,\tilde{x}_{T-1})$, $(x_0,\hdots,x_{T-1})$ and $(w_0,\hdots,w_{T-1})$ onto the $l$-th Fourier basis function.
Now, instead of estimating $x$ in the function space \eqref{eq:lfp_model}, we alternatively estimate the sequence of Fourier coefficients $\{X_l\}$ in the sequence space using \eqref{eq:GaussianSequence}.
Pinsker's theorem gives an (asymptotically) minimax-optimal estimator for the Gaussian sequence model provided that the Fourier coefficients satisfy some predefined criteria.
Let the Fourier coefficients $X_l$ live in an ellipsoid such that $\sum_la_l^2X_l^2\leq C$ where $a_1 = 0$, $a_{2m}=a_{2m + 1} = (2m)^{\alpha}$ with $\alpha>0$ denoting the smoothness parameter. 
The minimax-optimal estimator of $X_l$ is given by \cite{Johnstone2012GaussianE}
\begin{align}\label{eq:PinskerEstimator}
    {X}_l \approx \left(1 - \frac{a_l}{\mu}\right)_{+} \tilde{X}_l,\quad\mu>0,\quad l=1,2,\hdots.
\end{align}
The function $(\cdot)_+$ operates as a rectified linear unit (ReLU), i.e., $(\cdot)_+ = \max\{\cdot,0\}$.
We see that Pinsker's estimator shrinks the observations $\tilde{X}_l$ by an amount $ 1 - a_l/\mu$ if $a_l<\mu$; otherwise, it attenuates them to zero.
Thus, the optimal estimator \eqref{eq:PinskerEstimator} yields only a \emph{finite} number of $L$ (complex) Fourier coefficients that correspond to the lowest $L$ frequencies (including the DC), where $L$ is the largest integer such that $a_L<\mu$ and $a_{L+1}\geq\mu$.

\subsection{Deep Cross-Subject Mapping}\label{sec_sysmodel:CVAE}

To learn the cross-subject neural activity map $g(\cdot)$ between Subjects A and B, we use deep conditional generative models such as the \emph{conditional variational autoencoder (CVAE)} \cite{CVAE_NIPS2015}; in our investigations, we have found out that the CVAE is \emph{robust} against cross-subject non-stationarity even when trained with limited data. 


The complete statistical description of the features of Subject A and B is given by the joint distribution $p_\theta({X},{Y})$.\footnote{Recall that both $X$ and $Y$ correspond to the same action in $\mathcal{K}$ (we have dropped the class index $k$ to simplify the notation).}
Knowing $p_\theta({X},{Y})$ allows us to compute the transfer function as a \emph{multivariate regression function}; for the squared loss, this is simply the mean of the conditional distribution $p_\theta({Y}|{X})$:
\begin{align}\label{eq:transfer_as_regression}
    g({X}) = \mathbb{E}_{{Y}\sim p_\theta({Y}|{X})}[{Y}].
\end{align}

\begin{figure}
    \centering
    \includegraphics[scale = 0.325]{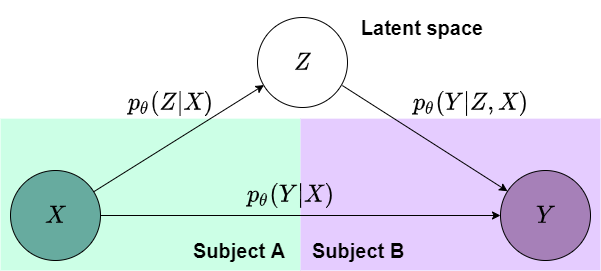}
    \caption{Graphical representation of the conditional variational autoencoder.}
    \label{CVAE_graph}
\end{figure}

To estimate the parameters $\theta$, we apply a CVAE approach, which is designed to model complex, multimodal output distributions by allowing the prior distribution of a low-dimensional \emph{latent} variable $Z\in\mathbb{R}^{M},M<D$ to be modulated by the input distribution, see Fig.~\ref{CVAE_graph}.
Given a feature vector ${X}$ from Subject A and following the generative process of the CVAE, the latent variable $Z$ is drawn from the conditional prior distribution $p_\theta(Z|{X})$; the representation of ${X}$ in the feature of Subject B is then generated according to $p_\theta({Y}|{X},Z)$, which is also known as the \emph{generation} model or \emph{decoder}.


Estimating the conditional distribution $p_\theta({Y}|{X})$ using the above generative model is difficult due to intractable posterior $p_\theta(Z|{X},{Y})$ which renders the log-likelihood $\ln p_\theta({Y}|{X})$ intractable.
This is circumvented through variational inference where the \emph{variational lower bound (ELBO)} of $\ln p_\theta({Y}|{X})$, denoted by $\mathcal{L}_{\theta,\phi}({X},{Y})$, is used as a surrogate objective function \cite{kingma2013autoencoding}.
To approximate the posterior $p_\theta(Z|{X},{Y})$, a tractable proposal distribution $q_\phi(Z|{X},{Y})$, referred to as the \emph{recognition} function or \emph{encoder}, is introduced.
The ELBO can then be written as \cite{CVAE_NIPS2015}
\begin{align}\nonumber
    \mathcal{L}_{\theta,\phi}({X},{Y}) & = 
     - KL\left(q_\phi(Z|{X},{Y})||p_\theta(Z|{X})\right) \\\label{eq:cELBO}
     & + \mathbb{E}_{Z\sim q_\phi(Z|{X},{Y})}\left[\ln p_\theta({Y}|{X},Z)\right], 
\end{align}
where $KL(\cdot||\cdot)$ denotes the Kullback-Liebler (KL) divergence.
Thus, instead of maximizing the log-likelihood with respect to (w.r.t.) $\theta$, we maximize the ELBO w.r.t. $\theta$ and $\phi$ jointly. 

After obtaining the conditional prior $p_\theta(Z|{X})$ and the generator $p_\theta({Y}|{X},Z)$, we can proceed to infer the representation of a given feature vector $\hat{X}$ of Subject A into the feature space of Subject B, directly from $p_\theta({Y},Z|{X})$ as follows: 
\begin{align}
    \hat{Y} = g(\hat{X}) = \mathbb{E}_{{Y}\sim p_\theta({Y}|\hat{X},\hat{Z})}[{Y}],\quad \hat{Z}=\arg\max_{Z}p_\theta(Z|\hat{X}).
\end{align}
Note that the CVAE introduces \emph{three} new conditional distributions: the prior $p_\theta(Z|{X})$, the generator $p_\theta({Y}|Z,{X})$ and the recognizer $q_\phi(Z|{X},{Y})$, all of which need to be learned in order to estimate the transfer function.
Thus, the introduction of a latent variable may seem disadvantageous at first; nevertheless, as our investigations have shown, this comes with the benefit of increased performance since it allows the CVAE to model complex, multimodal latent distributions.
Intuitively, we expect the recognizer and the prior to utilize knowledge from both subjects and learn to encode features corresponding to same tasks $k\in\mathcal{K}$ into the same region in the latent space in a subject-invariant manner.
As a result, Subject A data point that is initially difficult to decode in the Subject A feature space, might be easier to classify in the feature space of Subject B after reconstructing the representation from its latent encoding.     

\subsubsection{Comparison with other deterministic/non-deterministic methods} We have found that generative adversarial networks (GANs) can not be trained reliably on our setup due to limited data. Similarly, non-generative models such as MLPs and auto-encoders fail to perform well due to lack of robustness to the neural variability.
In our experience, the CVAE approach strikes the best trade-off between reliability, performance and robustness.

\subsection{Neural Decoder}\label{sec_sysmodel:MLPdecoders}

The decoding method is simply a classification model, trained on the feature space of subject B, that takes the feature vector from Subject X (transferred using the cross-subject map) and infers the task from the set of motor actions $\mathcal{K}$. Taking advantage of the non-parametric feature extraction method, one could also leverage a deep neural network for classification as in \cite{Angjelichinoski_2020_DJSNN}, yielding a fully deep end-to-end cross-subject neural decoding system.  


\section{Evaluations}\label{sec:eval}


\subsection{Training and Hyperparameters}\label{sec_eval:training}

LFP activity was acquired synchronously from each electrode with no delay w.r.t. the onset of the memory period (i.e., the sampling begins at the beginning of the memory period).
Pinsker's feature extraction is applied to each electrode separately, yielding $N$ feature vectors (one per electrode) with dimension $2L-1$; note that we use the real Cartesian coordinates to represent and process the complex Fourier coefficient.
These feature vectors are then concatenated horizontally to form one large feature vector of dimension $D=N\cdot(2L-1)$.

\begin{figure*}
\centering
\subfloat[Training]{\includegraphics[scale = 0.22]{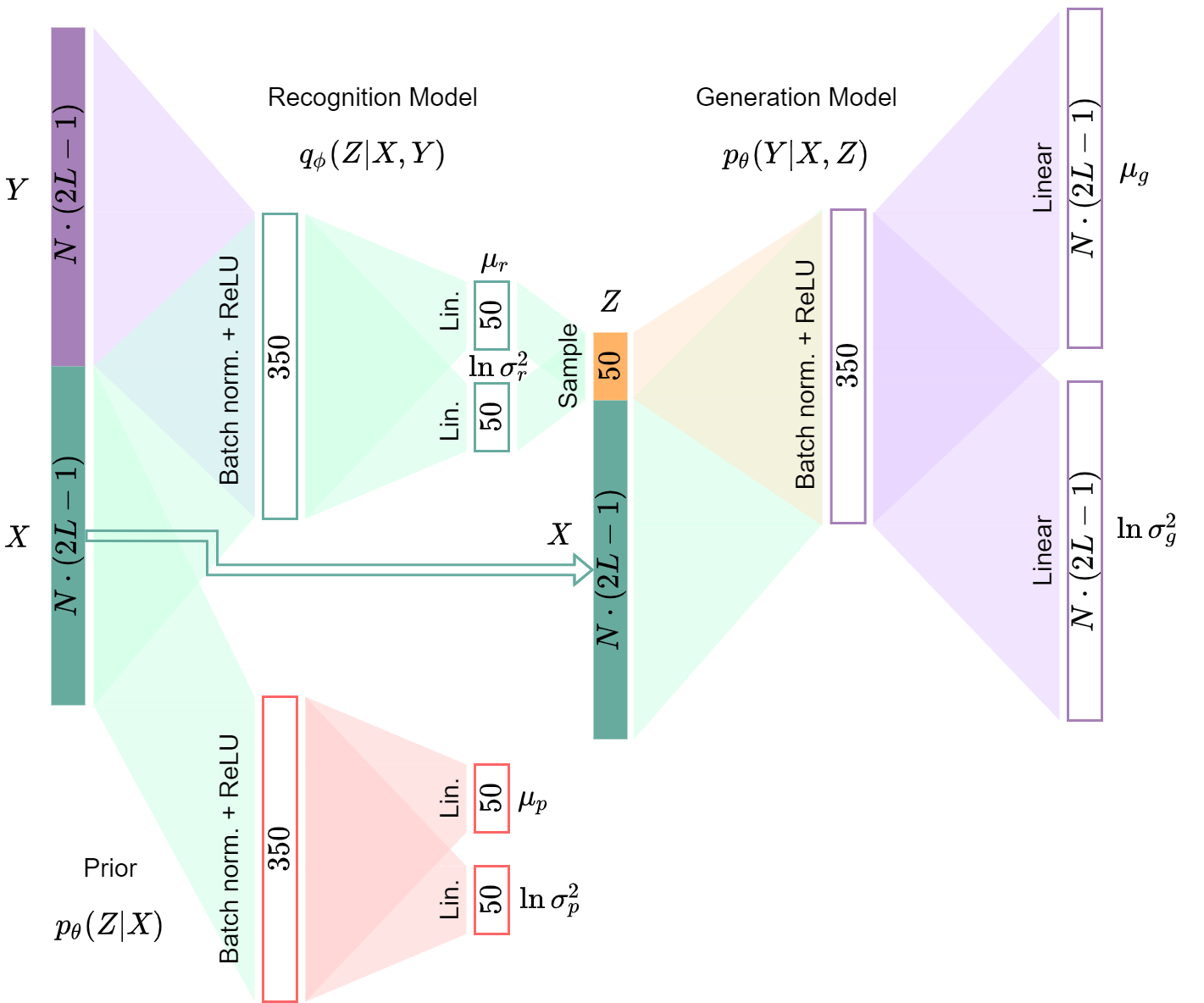}\label{CVAE_train}}
\hfil
\subfloat[Testing]{\includegraphics[scale = 0.22]{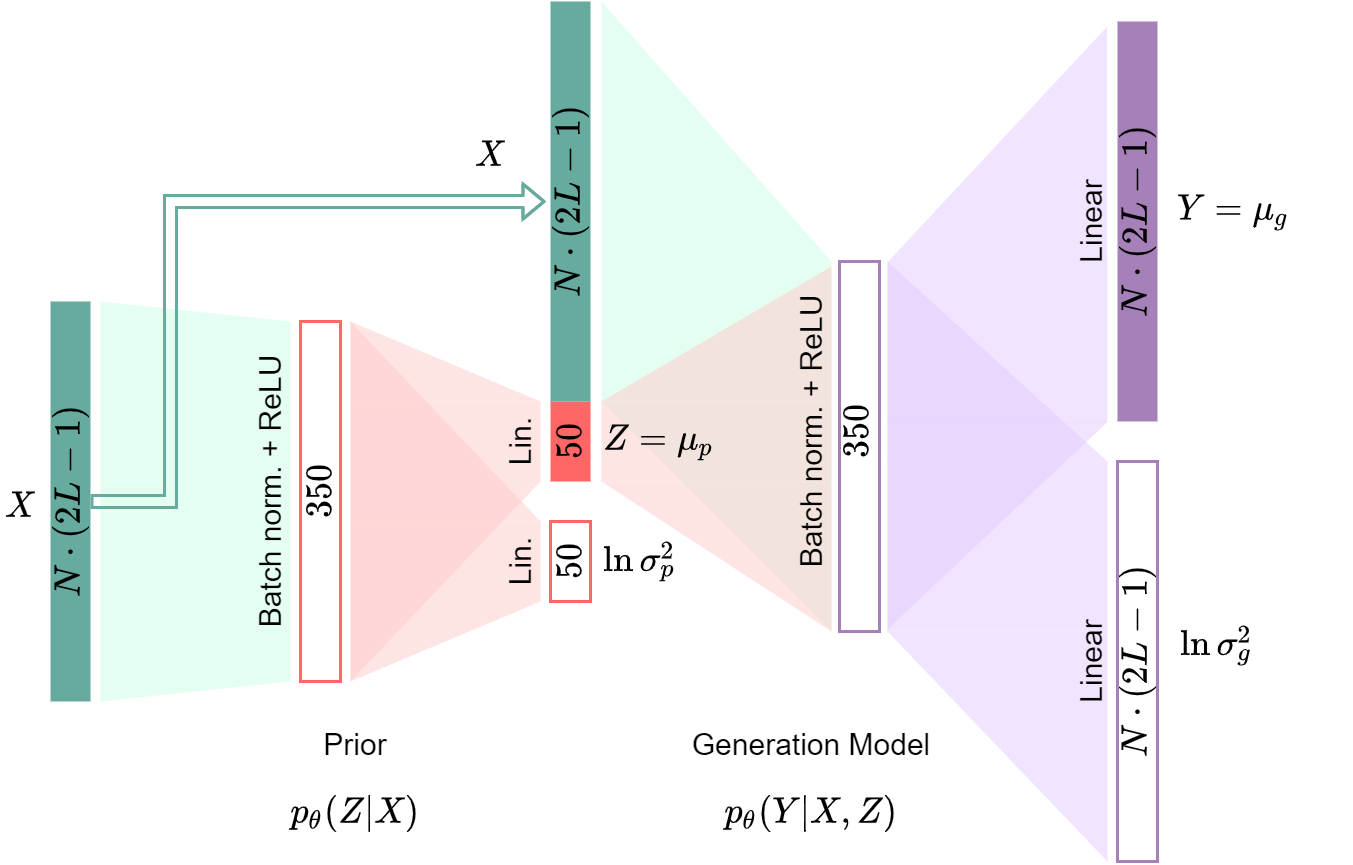}\label{CVAE_test}}
\caption{Schematic illustration of the MLP-based CVAE architecture.}
\label{CVAE_architecture}
\end{figure*}

For convenience, let $\mathcal{D}_\text{A}=\{{X}(i)\}$ and $\mathcal{D}_\text{B}=\{{Y}(i)\}$ denote the training data sets of Subjects A and B, where ${X}(i)$ and ${Y}(i)$ are the $i$-th feature vectors. 
It is a common practice to use neural networks to model the prior, recognition and generation distributions. 
Here, we adopt disentangled Gaussian distributions for the prior and the generator/recognizer: $p_\theta(Z|X)=\mathcal{N}(\mu_p,\text{diag}(\sigma_g^2))$, $q_\phi(Z|{X},{Y})=\mathcal{N}(\mu_r,\text{diag}(\sigma_r^2))$, $p_\phi({Y}|{X},Z)=\mathcal{N}(\mu_g,\text{diag}(\sigma_g^2))$; the mean and variance vectors are given with the outputs of deep MLP networks.
Using these models, the ELBO evaluates to
\begin{align}\nonumber
   \mathcal{L}_{\theta,\phi}({X},{Y}) & \approx \sum_{m=1}^M\left( 1 + \ln\frac{\sigma_{r,m}^2}{\sigma_{p,m}^2} - \frac{\sigma_{r,m}^2+(\mu_{r,m}-\mu_{p,m})^2}{\sigma_{p,m}^2}\right) \\\label{eq:cELBO_gaussian}
   & - \sum_{d=1}^D\left(\ln\sigma_{g,d}^2+ \frac{({Y}_d-\mu_{g,d})^2}{\sigma_{g,d}^2}\right),
\end{align}
where $\mu_g,\sigma_g$ are given as the outputs of the generation network, excited by $Z = \mu_r + \sigma_r\odot\epsilon$, $\epsilon\sim\mathcal{N}(0,I_M)$, whereas $\mu_r,\sigma_r$ and $\mu_p,\sigma_p$ are given by the outputs of the recognition and prior MLP networks, respectively, see Fig.~\ref{CVAE_train}; note that we used the one-sample estimate of the mean to approximate the expectation in the second term in \eqref{eq:cELBO} as well as the standard reparametrization trick to sample from $q_\phi(Z|{X},{Y})$; note that, as the output features in the above model are assumed to be independent, the training examples in $\mathcal{D}_\text{B}$ are first decorrelated before training the CVAE.

Minimizing the ELBO trains the CVAE to reconstruct the output ${Y}$ from an input ${X}$; if ${X}$ is drawn from the class-conditional distribution corresponding to task $k\in\mathcal{K}$, then ${Y}$ should be also drawn from the same class-conditional distribution in the feature space of Subject B.
Hence, the trials in the training data sets should be organized such that there is an input-output correspondence with respect to the task $k$.
Recall that the experimental protocol is randomized w.r.t. the sequence of targets which are selected uniformly and independently for each subject; as a result, we are unable to establish correspondence between the trails from each class-conditional distribution. 
One way to address this issue is to assign the correspondences between training trials from the same class-conditional distributions at \emph{random}. 
This might require large training data in order to average out the assignment randomness.
An alternative approach is to replace each trial in $\mathcal{D}_\text{B}$ with the corresponding class-conditional mean vector, which bears conceptual similarity with the approach undertaken in \cite{Angjelichinoski_2020_CS} based on the method of moments. 
Note that 
even though we seek to find a transfer function $g(\cdot)$ valid for all $k\in\mathcal{K}$, the task information will still be exploited implicitly during training, whereas the model remains unsupervised in testing time.

The classifier $\delta_\text{B}(\cdot)$ is trained on the same set of features $\mathcal{D}_\text{B}$ as the CVAE (more specifically, an extended data set with the corresponding task labels).
We opt for this approach due lack of sufficient data to allow the CVAE and the classifier to be trained independently. The approach is well suited to the non-stationary data, as in such cases, independent training data sets might correspond to different distributions which introduces a fundamental mismatch between the transfer function model and the feature space represented by the neural decoder.

Previous studies in \cite{Angjelichinoski2019} have shown that most of the information relevant for the decoding of eye movement intentions from LFPs is stored within the first half of the memory period; 
motivated by this, we fix $T=650$ to correspond roughly to the first half of the memory period. 
Further, we fix the dimension of the latent space to be $M=50$ and we use fully-connected MLPs with one hidden layer with $350$ neurons and ReLU activation functions for each of the conditional distributions. 
Fig.~\ref{CVAE_graph} shows the graphical representation of the CVAE.
The implementation of the graphical model using multilayer perceptrons (MLPs) to parameterize the recognition, generation and prior distributions is shown in Fig.~\ref{CVAE_architecture}.
All MLPs are shallow, with a single hidden layer. 
For numerical stability, we also applied batch normalization before applying the non-linear activation in each of the MLPs.
We trained the CVAE using Adadelta \cite{zeiler2012adadelta} optimization method.
In our implementation, the learning rate is adjusted in each training epoch. 
Specifically, let $\lambda_{i},i=1,2,\hdots$ denote the learning rate in the $i$-th training epoch; the learning rate in the following epoch is adjusted as
\begin{align}\label{eq:lr}
    \lambda_{i+1} = \gamma \lambda_{i},\quad i=1,2,\hdots,
\end{align}
where $\gamma\leq 1$ is a decay factor; in our implementation $\gamma = 0.99$.

The following CVAE parameters are treated as hyper-parameters over which the performance of the cross-subject neural decoder is optimized:
\begin{itemize}
    \item initial learning rate used in the first training epoch $\lambda_1$;
    \item number of hidden neurons in the recognizer $H_r$, generator $H_g$ and the prior $H_p$;
    \item the dimension of the latent space $M$;
    \item size of the minibatch
\end{itemize}
To determine suitable values for the above hyper-parameters, we conducted a grid-based cross-validation search in the hyper-parameters space, at superficial cortical depth; namely, we fixed $L=3$ and we used several random training/evaluation subset splits of size 1200/200 for the two data sets collected at $\approx$ 1 mm and $\approx$ 1.4 mm for Monkey B and A, respectively.
Throughout these evaluations, we have found out that the performance of the cross-subject neural decoder (in terms of cross-validation risk) is most susceptible to the learning rate $\lambda_1$ and (to a lesser extent) to the size of the minibatch and the dimension of the latent space $M$.
For instance, we have found out that the model becomes overfitted for learning rates larger than $0.2$ and undefitted for learning rates smaller than $0.05$; initial learning rates between $0.1$ and $0.15$ were found to give good convergence properties and we selected $\lambda_1 = 0.125$ for the evaluations presented in the manuscript.
In a similar manner, we selected $H_r=H_p=H_g=350$, $M=50$ and the size of the minibatch to be $75$.
We note that these preliminary evaluations are conducted over the data sets collected at superficial cortical sites in both subjects and were only used to gain coarse and loose sense about the intervals in which suitable values for the hyper-parameters can be found; a primary reason why we did not perform exhaustive fine-tuning of the hyper-parameters is the fact that 1) the LFP activity is non-stationary, and 2) the data is limited.
In other words, even if we optimize and fine-tune the hyper-parameter over a given randomly allocated training data set, due to the non-stationary nature of the LFP activity, that does not imply that the same hyper-parameter will be optimal for the test set.

Note that, during training the CVAE is trained to \emph{reconstruct} Subject B features $Y$ from Subject A features $X$ and $Z^{(s)}$ is sampled from the recognition model which encodes the information from both feature spaces; however, in testing time, we wish to \emph{infer} the Subject B feature vector $Y$ from a given Subject A feature vector $X$ and to do this we use the conditional prior $p_\theta(Z|X)$.
To balance these two tasks, namely reconstruction and inference, the authors in \cite{CVAE_NIPS2015} propose a modified optimization objective, i.e., ELBO that combines these two tasks by adding a term that accounts for the inference task, obtained by setting $q_\phi(Z|X,Y)=p_\theta(Z|X)$; nevertheless, in our investigations, the performance difference between the original and the modified ELBO formulations was statistically insignificant.

After training the parameters of the prior and the generator, we use the CVAE to map a given feature vector $X$ into a corresponding $Y$.
This can be done in a generative manner, following the generative process of the graph in Fig.~\ref{CVAE_graph}.
An alternative option is to use deterministic inference, without sampling, as in Fig.~\ref{CVAE_test}, following eq.(9) in the manuscript. We have tested both, the generative and the deterministic approach and they produce practically the same results in terms of convergence of the average decoding accuracy, which the generative approach yielding slightly larger variance of the decoding accuracy.

\begin{table*}
  \caption{\centering Cross-subject decoding accuracy (in \%). The subject-specific (\emph{local}) decoding accuracy of Monkey B (Subject B) at cortical depth of 1 mm is \textbf{87.8$\pm$1.2}\% with $L=3$.}
  \label{table:results_A_allS_L5}
  \centering
  \begin{tabular}{|l|l|l|l|l|}
    \toprule
    \multicolumn{1}{c|}{} & \multicolumn{4}{c|}{Mean cortical depth of Monkey S (Subject X)} \\
    \cmidrule(r){2-5}
    \multicolumn{1}{c|}{} & 1.4 mm & 2.2 mm & 2.8 mm & 3.5 mm \\
    \cmidrule(r){1-5}
    Subject X local ($L=$3) & 75.0$\pm$3.2 & 66.2$\pm$3.3 & 51.8$\pm$2.8  & 45.8$\pm$4.0 \\
    \cmidrule(r){1-5}
    Cross-subject, X$\rightarrow$Y direct ($L=$3) & 12.4$\pm$2.6 & 13.4$\pm$3.5 & 12.82$\pm$2.9  & 11.9$\pm$2.4 \\
    \cmidrule(r){1-5}
    Cross-subject, X$\rightarrow$Y linear map \cite{Angjelichinoski_2020_CS} & 55.0$\pm$0.0 & 40.2$\pm$0.0 & 36.0$\pm$0.0  & 35.4$\pm$0.0 \\
    \cmidrule(r){1-5}
    Cross-subject, X$\rightarrow$Y CVAE ($L=$3) & \textbf{81.0$\pm$3.0} & \textbf{72.2$\pm$2.8} & \textbf{57.0$\pm$2.5}  & \textbf{50.0$\pm$3.4} \\
    \cmidrule(r){1-5}
    Relative gain in \% ( w.r.t. linear map) & {46.3}$\pm$5.6 & {79.6}$\pm$7.0 & {57.1}$\pm$7.0 & {40.0}$\pm$9.7 \\
    \cmidrule(r){1-5}
    Relative gain in \% ( w.r.t. local) & {7.5}$\pm$3.6 & {9.2}$\pm$5.1 & {9.4}$\pm$6.6 & {8.7}$\pm$8.9 \\
    \bottomrule
  \end{tabular}
\end{table*}

\begin{table*}
  \caption{\centering Cross-subject decoding accuracy (in \%) at superficial cortical depth w.r.t. number of Fourier coefficients $L$ per channel.}
  \label{table:results_a_S1_allL}
  \centering
  \begin{tabular}{|l|l|l|l|l|l|}
    \toprule
    \multicolumn{1}{c|}{} & \multicolumn{2}{c|}{Subject-specific (local)} & \multicolumn{3}{c|}{Cross-subject, A$\rightarrow$B} \\
    \cmidrule(r){2-6}
    \multicolumn{1}{c|}{} & Mon.A(1.4 mm) & Mon.B(1 mm) & Direct & CVAE & Rel.gain (local) \\
    \cmidrule(r){1-6}
    $L=2$ & 72.2$\pm$3.0 & 85.9$\pm$1.4 & 13.6$\pm$2.7 & \textbf{78.0$\pm$2.1} & 7.2$\pm$4.3 \\
    \cmidrule(r){1-6}
    $L=3$ & 75.0$\pm$3.2 & 87.8$\pm$1.2 & 12.4$\pm$2.6 & \textbf{81.0$\pm$3.0} & 7.5$\pm$3.6 \\
    \cmidrule(r){1-6}
    $L=4$ & 75.0$\pm$2.5 & 88.1$\pm$1.2 & 11.9$\pm$2.8 & \textbf{80.1$\pm$2.0} & 6.8$\pm$3.9 \\
    \cmidrule(r){1-6}
    $L=5$ & 74.7$\pm$2.6 & 87.2$\pm$1.1 & 12.5$\pm$2.0 & \textbf{79.0$\pm$2.5} & 5.4$\pm$3.4 \\
    \cmidrule(r){1-6}
    $L=6$ & 73.2$\pm$3.1 & 89.0$\pm$1.0 & 13.2$\pm$3.1 & \textbf{75.0$\pm$2.4} & 2.0$\pm$4.0 \\
    \cmidrule(r){1-6}
    $L=7$ & 71.8$\pm$2.9 & 88.0$\pm$1.0 & 13.6$\pm$2.8 & \textbf{73.0$\pm$3.2} & 1.1$\pm$4.4 \\
    \bottomrule
  \end{tabular}
\end{table*}

\subsection{Results}\label{sec_eval:results}

To avoid over-fitting and obtain reliable conclusions, we apply statistical averaging where the performance metric, namely the \emph{average decoding accuracy}, is computed as an empirical average over multiple randomly selected testing data sets.
The staring data sets, each of them containing $1400$ trials are randomly split into two \emph{disjoint} subsets with sizes $1200$ for training and $200$ trials for testing.
The neural decoder is then trained using the training subset and the decoding accuracy over the test subset is recorded.  
This procedure is repeated $100$ times, and the final estimate of the decoding accuracy is computed by averaging the individual decoding accuracy from each round.

Table~\ref{table:results_A_allS_L5} presents the results from the evaluations where we fixed $L=3$, corresponding to a cut-off frequency of $\approx 4$ Hz. 
The data set from Monkey B results in a more reliable subject-specific, i.e., \emph{local} neural decoder in comparison with the Monkey A, even at neurobiologically similar cortical sites, near the surface of the PFC; the same was observed in \cite{Markowitz18412,Angjelichinoski2019}.
After mapping Monkey A (test) data onto the feature space (spanned by the training portion) of Monkey B using the CVAE approach, we observe that the cross-subject decoder consistently \emph{outperforms} the local Monkey A neural decoder across all cortical depths.
This result confirms the remarkable robustness of the CVAE against the non-stationary nature of LFP activity across subject, and is further strengthened when compared against the direct cross-subject decoding, without neural mapping (whose performance resembles the performance of a random choice decoder), as well as the cross-subject decoding with linear transfer functions from \cite{Angjelichinoski_2020_CS}.
We note that the linear map results are taken directly from \cite{Angjelichinoski_2020_CS} for the purpose of qualitative comparison even if the evaluation methodology used there, including the allocation of training/testing subsets differs slightly.

\begin{table*}
  \caption{\centering Cross-subject decoding accuracy (in \%). The subject-specific (\emph{local}) training/testing decoding accuracy of Monkey B (Subject B) at cortical depth of 1 mm is \textbf{100.0$\pm$0.0}\%/\textbf{87.8$\pm$1.2}\%. All results are obtained for $L=$3. 
  }
  \label{table:results_A_allS_L5_II}
  \centering
  \begin{tabular}{|l|l|l|l|l|}
    \toprule
    \multicolumn{1}{c|}{} & \multicolumn{4}{c|}{Mean cortical depth} \\
    \cmidrule(r){2-5}
    \multicolumn{1}{c|}{} & 1.4 mm & 2.2 mm & 2.8 mm & 3.5 mm \\
    \cmidrule(r){1-5}
    Monkey A (Subject A) local: Training & 100.0$\pm$0.0 & 99.9$\pm$0.1 & 99.7$\pm$0.3  & 99.9$\pm$0.1 \\
    \cmidrule(r){1-5}
    Monkey A (Subject A) local: Testing & 75.0$\pm$3.2 & 66.2$\pm$3.3 & 51.8$\pm$2.8  & 45.8$\pm$4.0 \\
    \cmidrule(r){1-5}
    Cross-subject, A$\rightarrow$B CVAE: Training & {98.9$\pm$0.3} & {97.4$\pm$0.5} & {90.0$\pm$0.7}  & {95.0$\pm$0.9} \\
    \cmidrule(r){1-5}
    Cross-subject, A$\rightarrow$B CVAE: Testing & {81.0$\pm$3.0} & {72.2$\pm$2.8} & {57.0$\pm$2.5}  & {50.0$\pm$3.4} \\
    \bottomrule
  \end{tabular}
\end{table*}

\begin{table*}
  \caption{\centering Cross-subject decoding accuracy (in \%) at superficial cortical depth w.r.t. number of complex Fourier coefficients $L$ per channel. The average training accuracy of the local decoders for both Monkey A (1.4 mm) and B (1 mm) converged to 100.0\%.}
  \label{table:results_a_S1_allL_II}
  \centering
  \begin{tabular}{|l|l|l|l|l|l|}
    \toprule
    \multicolumn{1}{c|}{} & \multicolumn{2}{c|}{Subject-specific (local)} & \multicolumn{2}{c|}{Cross-subject, A$\rightarrow$B} \\
    \cmidrule(r){2-5}
    \multicolumn{1}{c|}{} & Mon.A(1.4 mm) & Mon.B(1 mm) & CVAE: Testing & CVAE: Training \\
    \cmidrule(r){1-5}
    $L=2$ & 72.2$\pm$3.0 & 85.9$\pm$1.4 & \textbf{78.0$\pm$2.1} & 94.5$\pm$0.7 \\
    \cmidrule(r){1-5}
    $L=3$ & 75.0$\pm$3.2 & 87.8$\pm$1.2 & \textbf{81.0$\pm$3.0} & 98.9$\pm$0.3 \\
    \cmidrule(r){1-5}
    $L=4$ & 75.0$\pm$2.5 & 88.1$\pm$1.2 & \textbf{80.1$\pm$2.0} & 99.7$\pm$0.1 \\
    \cmidrule(r){1-5}
    $L=5$ & 74.7$\pm$2.6 & 87.2$\pm$1.1 & \textbf{79.0$\pm$2.5} & 99.8$\pm$0.1 \\
    \cmidrule(r){1-5}
    $L=6$ & 73.2$\pm$3.1 & 89.0$\pm$1.0 & \textbf{75.0$\pm$2.4} & 99.9$\pm$0.0 \\
    \cmidrule(r){1-5}
    $L=7$ & 71.8$\pm$2.9 & 88.0$\pm$1.0 & \textbf{73.0$\pm$3.2} & 100.0$\pm$0.0 \\
    \bottomrule
  \end{tabular}
\end{table*}

\begin{table*}
  \caption{\centering Cross-subject decoding accuracy (in \%) at superficial cortical depth w.r.t. number of complex Fourier coefficients $L$ per channel. The average training accuracy of the local decoders for both Monkey A (2.2 mm) and B (1 mm) converged to 100.0\%.}
  \label{table:results_a_S2_allL_II}
  \centering
  \begin{tabular}{|l|l|l|l|l|l|}
    \toprule
    \multicolumn{1}{c|}{} & \multicolumn{2}{c|}{Subject-specific (local)} & \multicolumn{2}{c|}{Cross-subject, A$\rightarrow$B} \\
    \cmidrule(r){2-5}
    \multicolumn{1}{c|}{} & Mon.A(2.2 mm) & Mon.B(1 mm) & CVAE: Testing & CVAE: Training \\
    \cmidrule(r){1-5}
    $L=2$ & 62.0$\pm$2.8 & 85.9$\pm$1.4 & \textbf{68.2$\pm$3.0} & 90.5$\pm$0.8 \\
    \cmidrule(r){1-5}
    $L=3$ & 66.2$\pm$3.3 & 87.8$\pm$1.2 & \textbf{72.2$\pm$2.8} & 97.4$\pm$0.5 \\
    \cmidrule(r){1-5}
    $L=4$ & 66.3$\pm$3.2 & 88.1$\pm$1.2 & \textbf{72.0$\pm$3.0} & 98.8$\pm$0.3 \\
    \cmidrule(r){1-5}
    $L=5$ & 66.6$\pm$3.0 & 87.2$\pm$1.1 & \textbf{70.3$\pm$3.3} & 99.7$\pm$0.1 \\
    \cmidrule(r){1-5}
    $L=6$ & 65.5$\pm$3.3 & 89.0$\pm$1.0 & \textbf{68.1$\pm$3.6} & 99.9$\pm$0.1 \\
    \cmidrule(r){1-5}
    $L=7$ & 64.2$\pm$3.7 & 88.0$\pm$1.0 & \textbf{64.5$\pm$3.3} & 99.9$\pm$0.0 \\
    \bottomrule
  \end{tabular}
\end{table*}

Table~\ref{table:results_a_S1_allL} shows the result from the cross-subject neural decoding at superficial cortical depths, for a range of cut-off frequencies, i.e., number of retained (complex) Fourier coefficients $L$ per channel.
We observe that the cross-subject decoding accuracy reaches its peak of $79-81\%$ for cut-off frequencies between $2$ and $7$ Hz ($L$ between $2$ and $5$), after which it begins to deteriorate; in fact, for $L\geq 8$, the subject-specific, local decoder of Monkey A dominates the cross-subject decoding performance.
This is an interesting result that confirms and further extends previous findings that the information pertinent to the neural decoding of memory-guided eye movement goals is available within the lowest band of the LFP frequency spectrum, between 0 and 10 Hz \cite{Angjelichinoski2019}.

Tables~\ref{table:results_A_allS_L5_II} and ~\ref{table:results_a_S1_allL_II} complement Tables~\ref{table:results_A_allS_L5} and ~\ref{table:results_a_S1_allL} by showing the average decoding accuracy over the training subsets, in addition to the testing decoding accuracy.
Since the values of the hyper-parameters were optimized over the data sets collected at superficial cortical sites in both subjects, the cross-subject training accuracy might suggest that the selected values are not optimal for the data sets collected at deeper cortical sites. 
In other words, the cross-subject decoding performance might benefit from further fine-tuning of the CVAE hyper-parameters, including the (initial) learning rate, for each cortical site and each $L$ individually.
Some preliminary investigations that we have conducted demonstrate that it is indeed possible to slightly improve the results at deep cortical sites and for $L>7$ by further fine-tuning the learning rate; nevertheless, this improvement has so far been only marginal.
We conclude that further investigation is required which is outside the scope of the work presented in this paper.

Table~\ref{table:results_a_S2_allL_II} is similar to Table~\ref{table:results_a_S1_allL} and  shows the cross-subject decoding training/testing performance at (superficial) cortical depth of 2.2 mm for Monkey A.
We observe the same trend as in Table~\ref{table:results_a_S1_allL}; namely, the cross-subject decoder outperforms the local, subject-specific decoder of Monkey A within the lowest 10 Hz of the LFP frequency band after which is begins to deteriorate and is dominated by the performance of the subject-specific decoder of Monkey A.

\begin{figure*}
\centering
\subfloat[]{\includegraphics[scale=0.5]{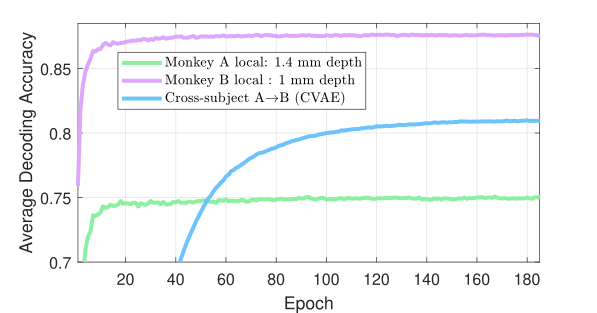}\label{results_L3_a}}
\hfil
\subfloat[]{\includegraphics[scale=0.5]{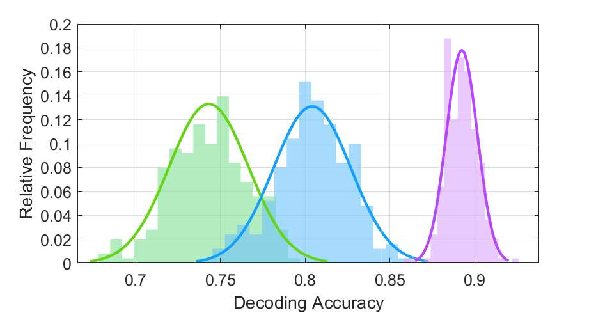}\label{results_L3_b}}
\caption{First and second order analysis of the cross-subject decoding accuracy at superficial cortical depth with $L=3$ complex Fourier coefficients per electrode.}
\label{results_L3}
\end{figure*}

Fig.~\ref{results_L3} expands on the results in Table~\ref{table:results_a_S1_allL} and shows the results for $L=3$ in more detail where the Fig.~\ref{results_L3_a} shows the convergence of the average test decoding accuracy though the training epochs and Fig.~\ref{results_L3_b} shows the histograms of the local and cross-subject decoding accuracy.
The results in these plots demonstrate the shift of the cross-subject decoding distribution to the right by a relative margin of $\approx 8\%$ w.r.t. the subject-specific decoding distribution.

\begin{figure*}
    \centering
    \includegraphics[scale = 0.55]{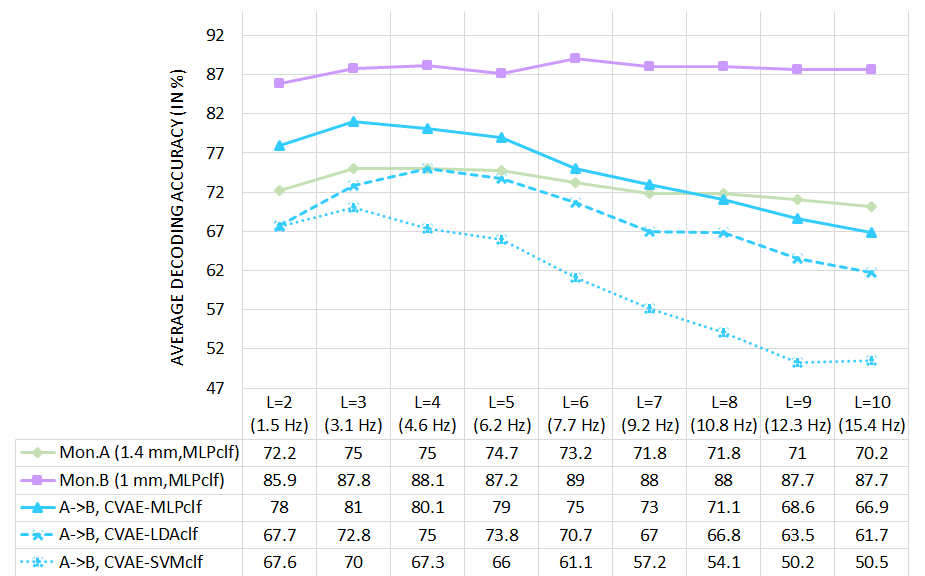}
    \caption{Cross-subject decoding at superficial cortical depth w.r.t. cut-off frequency: Comparison with linear neural decoders.}
    \label{CrossSubject_chart}
\end{figure*}

Fig.~\ref{CrossSubject_chart} shows an alternative representation of Table~\ref{table:results_a_S1_allL} with few additional results.
Specifically, it show the subject-specific and cross-subject neural decoding performances for $L>7$, i.e., cut-off frequencies larger than 10 Hz; this results confirm that for $L>7$, the neural decoding performance is becoming dominated by the local, subject-specific performance and there is no performance gain from cross-subject mapping. 
However, we observe that the cross-subject decoding even in this regime remains well above a random choice decoder, confirming once again the remarkable robustness of the CVAE.
Fig.~\ref{CrossSubject_chart} also shows the performance of popular linear classifiers such as linear discriminant analysis (LDA) and  multiclass support vector machine (SVM) for cross-subject neural decoding; this set of evaluations are also important since linear methods are a popular neural decoding method in the cross-subject literature due to limited training data \cite{Markowitz18412,Angjelichinoski2019}.
We observe that cross-subject decoding using LDA or SVM classifier performs well w.r.t. random choice.
However, they both fail to outperform the cross-subject MLP-based neural decoder as well as the MLP-based subject-specific decoder for Monkey A.
This is also expected; previous subject-specific investigations over the same experimental data \cite{Angjelichinoski_2020_DJSNN} have shown that MLP classifier outperforms LDA classifiers when trained over Pinsker's features.
As also shown, the SVM classifier performs comparatively worse than the LDA which also verifies previous findings \cite{Angjelichinoski2019} that the LDA classifier demonstrates the highest reliability and robustness among other popular linear classifiers such as SVM or even logistic regression. 

\section{Conclusions}

In this paper we considered the problem of mapping  neural activity from one Subject A to Subject B.
To address the issue of non-stationarity of neural activity, we proposed a solution based on deep generative model, i.e., conditional variational autoencoder.
We verified the viability of the proposed solution in the context of cross-subject neural decoding of motor intentions from local field potentials using a standard experiment in which two macaque monkeys perform memory-guided visual saccades to one of eight target locations on a screen.
The results demonstrate that the proposed cross-subject neural decoder 1) outperforms the local, subject-specific neural decoder of Subject A, by a relative margin of up to $10\%$, and 2) outperforms earlier benchmark based on linear cross-subject mapping \cite{Angjelichinoski_2020_CS} by a relative margin of up to $80\%$.

The findings we report have potentially far-reaching practical implications for the development of cross-subject BCIs in the domains of healthcare and public safety. 
Apart from civilian applications, BCIs are also foreseen as powerful emerging technological tool in the tactical domain. Even though our approach is application-agnostic, we note that the technological components of the proposed system are also applicable in scenarios of potential importance to national security.
We note that further investigations are required to generalize our results across different subjects and across a range of motor tasks.
Additional preclinical studies in animal models as well as clinical studies in patients are needed; this is a non-trivial, time-consuming and expensive endeavour.

\bibliographystyle{IEEEtran}
\bibliography{bare_jrnl}

\end{document}